\documentclass[letterpaper, 10 pt, journal, twoside]{ieeetran}
\IEEEoverridecommandlockouts
\usepackage{float}
\usepackage{graphicx}
\usepackage{amssymb}

\usepackage{amsmath}
\usepackage{mathtools}  
\usepackage{algorithmic}
\usepackage{algorithm}
\usepackage{nccmath}

\usepackage{amsthm}
\usepackage[usenames,dvipsnames,svgnames,table]{xcolor}
\theoremstyle{plain}

\newtheorem{definition}{Definition}

\usepackage{float}
\floatstyle{plaintop}
\restylefloat{table}
\usepackage{MnSymbol}%
\usepackage{wasysym}%
\usepackage{caption}
\usepackage{subcaption}

\usepackage{hyperref}

\renewcommand{\baselinestretch}{0.94}

\definecolor{deepblue}{rgb}{0,0,1}

\newcommand{\modification}[1]{{#1}}
\newcommand{\modificationtwo}[1]{{#1}}

\begin{document}

\title{ 
The Catenary Robot: Design and Control of a Cable Propelled by Two Quadrotors
}

\author{Diego S. D'antonio, Gustavo A. Cardona, and David Salda\~{n}a 
\thanks{Manuscript received: October, 15, 2020; Revised December, 13, 2020; Accepted February, 8, 2021.}
\thanks{This paper was recommended for publication by Editor Pauline Pounds upon evaluation of the Associate Editor and Reviewers' comments.}

\thanks{D. Salazar-D'antonio, G. A. Cardona, and D. Salda\~{n}a are with the Autonomous and Intelligent Robotics Laboratory (AIRLab) at Lehigh University, PA, USA:$\{$\texttt{diego.s.dantonio, gcardona, saldana\}@lehigh.edu}}   \thanks{Digital Object Identifier (DOI): see top of this page.}     

}     

\markboth{IEEE ROBOTICS AND AUTOMATION LETTERS. PREPRINT VERSION. ACCEPTED FEBRUARY, 2021}
{S. D'ANTONIO \MakeLowercase{\textit{et al.}}: THE CATENARY ROBOT: DESIGN AND CONTROL OF A CABLE PROPELLED BY TWO QUADROTORS} 

\maketitle

\begin{abstract}
Transporting objects using aerial robots has been widely studied in the literature. Still, those approaches always assume that the connection between the quadrotor and the load is made in a previous stage. However, that previous stage usually requires human intervention, and autonomous procedures to locate and attach the object are not considered. Additionally, most of the approaches assume cables as rigid links, but manipulating cables requires considering the state when the cables are hanging. In this work, we design and control a \emph{catenary robot}.  Our robot is able to transport hook-shaped objects in the environment. The robotic system is composed of two quadrotors attached to the two ends of a cable. By defining the catenary curve with five degrees of freedom, position in 3-D, orientation in the z-axis, and span, we can drive the two quadrotors to track a given trajectory. We validate our approach with simulations and real robots. We present four different scenarios of experiments. Our numerical solution is computationally fast and can be executed in real-time.

\end{abstract}


\IEEEpeerreviewmaketitle

\section{Introduction}


\IEEEPARstart{I}{n} recent years, aerial robots have become popular in industry and academia because of their low cost and a large number of applications. Especially in object transportation, aerial vehicles such as quadrotors have demonstrated to be effective as a solution.
Some of the main applications of quadrotors are:
delivering last-mile packages for retail and wholesale companies \cite{bamburry2015drones}; transporting supplies to disaster areas \cite{cacace2016control}; accessing dangerous areas for humans such as forest fires \cite{kinaneva2019early}, and delivering medicines and food such as needed in remote regions.


Quadrotors can overcome payload restrictions by cooperating with others to manipulate and transport objects that are either suspended or attached. 
In the aerial robotics literature, there is a large number of works that tackle the problem of cable-suspended load transportation, using either one quadrotor \cite{sreenath2013trajectory, cruz2017cable} or multiple \cite{Nathan2009, Sreenath2013DynamicsCA, cardona2019cooperative}.
\modification{In the case of multiple quadrotors, there are two well-known approaches, the point-mass load, and multiple contact points. In the point mass approach all the cables go to the same contact point in the mass \cite{cardona2019cooperative}. In the multiple contact approach, the robots are attached to different places on the load, which makes the problem more challenging since it involves rotational dynamics in addition. On the other hand, rigid cables \cite{Sreenath2013DynamicsCA} and flexible cables \cite{Kotaru2018}.  
With rigid cables, quadrotors are more susceptible to disturbances generated by other linked quadrotors. However, it is easier to localize where the load is due to the length is fixed. Flexible cables allow stretching, thus small disturbances do not affect the performance of linked  quadrotors too much. Nevertheless, localizing the load might suffer errors caused by vibrations. }
\begin{figure}[t]
    \centering
    {\includegraphics[width=0.85\linewidth, trim={0 1.1cm 0 1.2cm},clip]{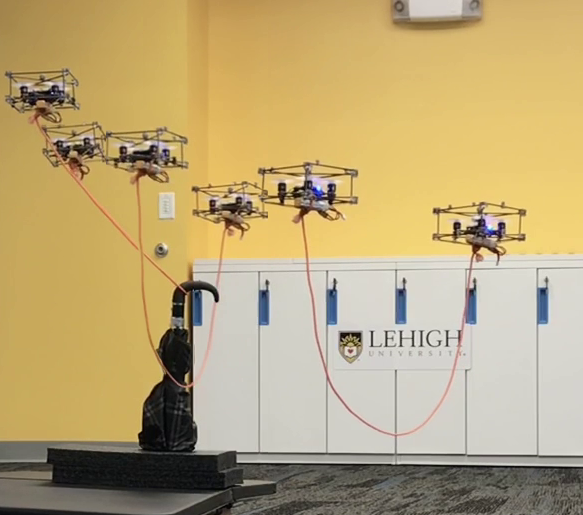}
    \caption{The \emph{catenary robot}, composed of a pair of quadrotors attached to the two ends of a cable.
    Controlling the lowest point, span and orientation of the curve, the \emph{catenary robot} can be used to interact and pull objects, in this case, an umbrella. 
    {Video available at:} \texttt{\href{https://youtu.be/3SaKKjl6os0}{https://youtu.be/3SaKKjl6os0}}
    }
    \label{fig:Introreal}}
\end{figure}

\begin{figure}[t]
    \centering
    {\includegraphics[width=0.4\textwidth]{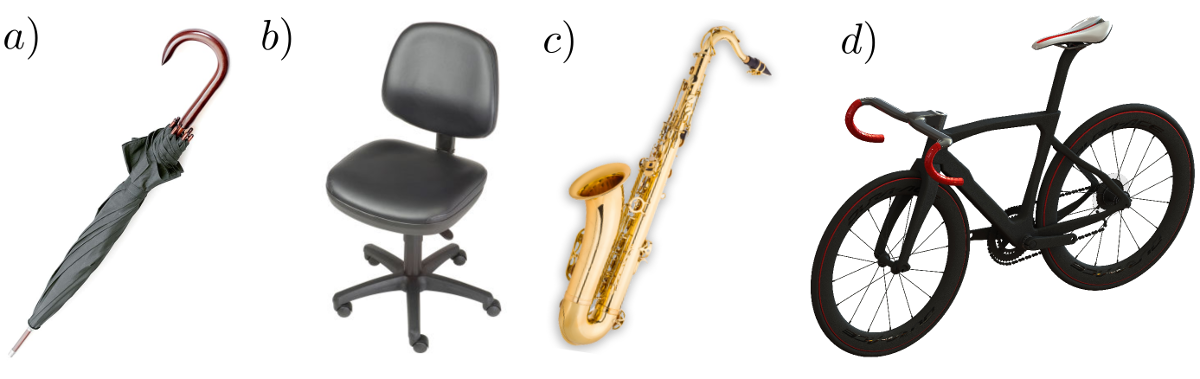}
    \caption{Objects that naturally have one or multiple hooks in their shape. a) an umbrella has a hook shape in its handle. b) A chair has narrow parts allowing a cable to go through them. c) A saxophone has a hook shape in its neck. d) A bicycle can be pulled passing a cable  through its handlebar or its saddle.}
    \label{fig:objects}}
\end{figure}

\modification{Despite the work} that has been done extensively in the suspended load transportation field, 
all the approaches assume that the connection between the quadrotor and the load is made in a previous stage.
However, that previous stage usually requires human intervention, and autonomous procedures to locate and attach the object are not considered.
Additionally, most of the approaches assume cables are rigid links, 
but manipulating cables 
\modification{requires to consider the case} 
when the cables are hanging.
In this work, we design and control a \emph{catenary robot} as in \modification{Fig.} \ref{fig:Introreal}, where a cable is attached to two quadrotors, allowing it to take its natural form caused by gravity and describing a catenary curve \cite{lockwood1967book}.
Catenary dynamics has been studied before \cite{Laranjeira2017}, including applications in robotics such as obstacle avoidance with a hanging cable or servoing visual approaches \cite{Galea2017, Xiao2018, Abiko2017}. There are also applications of lifting a hose with multiple quadrotors \cite{Kotaru2020}. 
\modification{However, those approaches were not developed with cable manipulation in mind.}
It is crucial to find a reference point for the catenary 
to attach to objects and manipulate them. We choose the lowest point of the curve as the reference because it can be used to attach objects. In this work, we consider objects that have embedded hooks in their shape, making them easy to be pulled without knotting or tightening.
Some examples of these objects are shown in Fig. \ref{fig:objects}.

The main contributions of this paper are twofold. First, we propose a versatile robot, called catenary robot, that is able to navigate in the environment and to get attached to objects with a hook-like shape. Second, we develop a trajectory-tracking method that controls the pose and the shape of the catenary curve. In the state of the art, the cable is modeled as a chain based on many links and joints but controlling \modification{such a mechanism} 
is highly expensive since each join increases the dimensions of the configuration space. In contrast, our method offers a fast numerical solution. 

\section{The Catenary Robot}
\label{Sec:Problem}
Cables offer a versatile way to interact and transport objects, but usually
it is challenging to find a way to attach cables to objects in an autonomous way.
In this work, we want to manipulate a cable using quadrotors in a way that it can be attached to hooks or objects with hook shapes, some of these objects can be seen in \modification{Fig.}~\ref{fig:objects}.
\modification{
Once the cable is attached to the object, other methods in the literature can be used for transportation \cite{Nathan2009, Sreenath2013DynamicsCA, cardona2019cooperative}.
}
The cable, hanging from the two quadrotors, adopts the form of the catenary curve, which varies depending on the position of the quadrotors, mostly in their altitude difference.
This lead us to the problem of finding a way to control the pose of the lowest point of the cable and the shape of the catenary curve.
Our robot is defined as follows.

\begin{definition}[\textbf{Catenary robot}]
A \emph{catenary robot} or \emph{flying catenary} is a
mechanical system composed of a cable with two quadrotors attached to its ends. Note that when the cable is hanging from the two quadrotors and there are no objects colliding with the cable, this naturally adopts a catenary curve form. The cable  is flexible and non-stretchable.


\end{definition}

\begin{figure}[t]
    \centering
    {\includegraphics[width=0.4\textwidth,height=5cm]{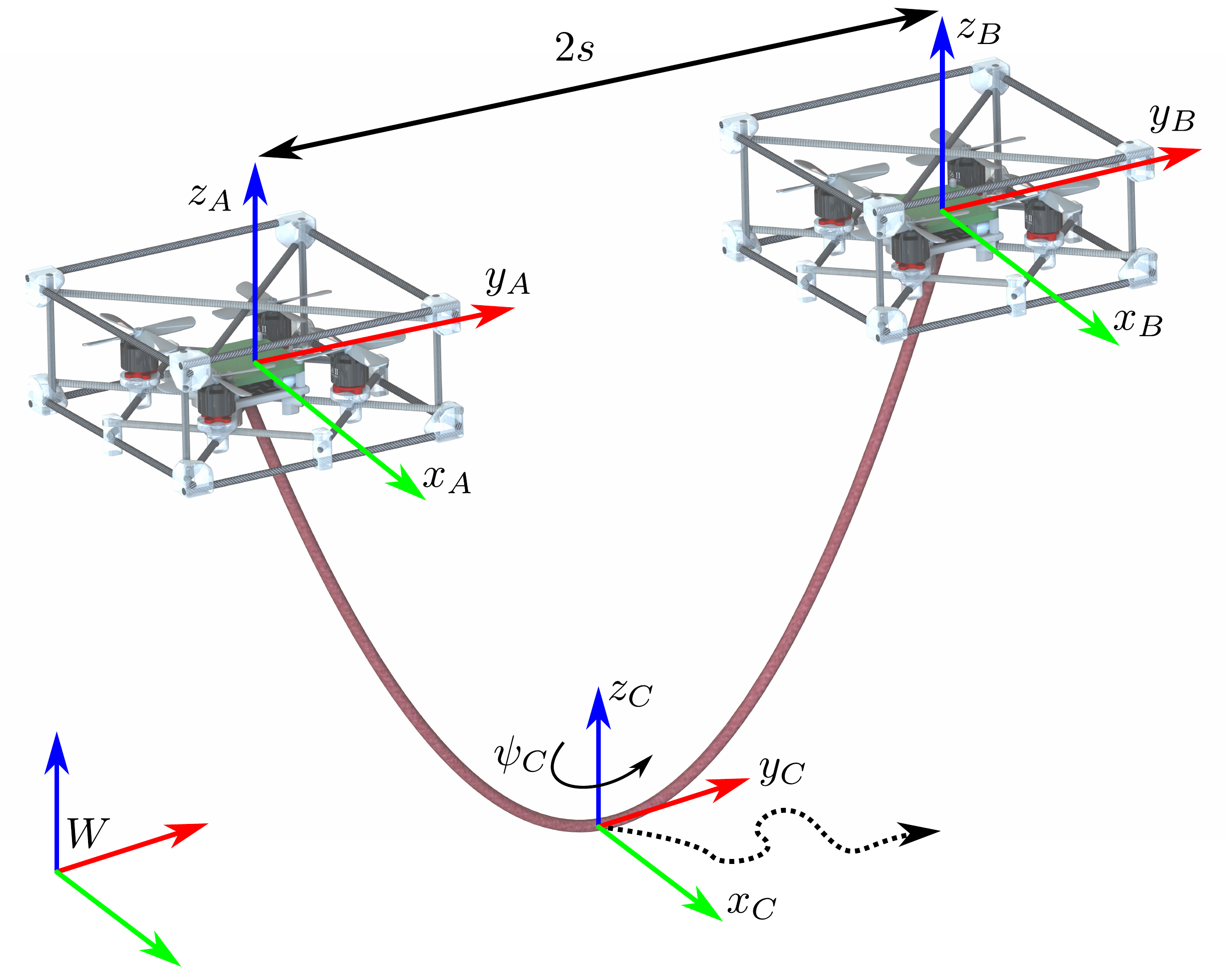}
    \caption{Coordinate frames of a catenary robot. 
    } 
    \label{fig:Intro}}
\end{figure}

The \emph{catenary robot} is illustrated in \modification{Fig.} \ref{fig:Intro}.
The {World frame, $\{\mathcal{W}\}$,} is a fixed coordinate frame with its $z$-axis pointing upwards. 
The reference point is the lowest point of the catenary in $\{\mathcal{W}\}$, denoted by $\mathbf{x}_C\in \mathbb{R}^3$.
The quadrotors are indexed as $A$ and $B$, and its locations in $\{W\}$ are denoted by $\mathbf{x}_A\in \mathbb{R}^3$ and $\mathbf{x}_B\in \mathbb{R}^3$, respectively.
\modification{
We assume that the quadrotors are always at the same altitude. In the case of perturbations, the controller has to drive them back to the same high.
}
\modification{
 Both quadrotors have same mass $m$ and inertia tensor~$\mathbf{J}$. The cable has a length $\ell$ and a mass $m_C$.
}
Each quadrotor has a frame, denoted by $\{\mathcal{A}\}$ and $\{\mathcal{B}\}$; the origin of each frame is located at its center of mass, the $x$-axis points to the front of the robot, and the $z$-axis points in the direction of the rotors. The orientation of quadrotors A and B with respect to the world frame are described by the rotation matrices $^W\mathbf{R}_A$ and $^W\mathbf{R}_B$ in SO(3), respectively.
\modification{
 The desired $x$-axis of each quadrotor is parallel to the $x$-axis of the catenary frame.
}

The catenary frame $\{\mathcal{C}\}$, has its origin at the lowest point of the catenary curve.
Assuming that the catenary is always on a vertical plane, called the catenary plane. The $x$-axis of $\{\mathcal{C}\}$ is  normal  to the catenary plane, and $y$-axis of $\{\mathcal{C}\}$ is defined by the unitarian tangent vector of the catenary curve at its lowest point.
The orientation of the catenary frame with respect to $\{\mathcal{W}\}$ is described by the rotation matrix $^W\mathbf{R}_C$.
In this paper, we assume that the robot moves slow enough such that the cable does not swing. Therefore, the catenary can only rotate with respect to its $z$-axis.
We denote the yaw angle of the catenary by $\psi_C$, and therefore its rotation matrix is  $^W\mathbf{R}_C= \text{Rot}_Z(\psi_C)$, where $\text{Rot}_Z$ is a function that returns the rotation matrix of an angle in the $z$-axis.
The quadrotors can keep a static orientation for the yaw angle but rotations around the center of the reference point can create torsion on the cable. Therefore, we want the yaw angle of the quadrotors to be the same as the yaw angle of the catenary.

The cable hangs from its two end points forming a catenary curve \cite{lockwood1967book}, starting at point~$\mathbf{x}_A$ and ending at point~$\mathbf{x}_B$. The equation of the curve in the catenary frame is  
 \begin{equation}
 \boldsymbol{\alpha}(r) = 
            \left\lbrack 
            \begin{array}{c}
                \modification{0}\\
                \modification{r} \\
                a \left ( \cosh \frac{{r}}{a} - 1 \right )
            \end{array}
            \right\rbrack,
            \label{eq:catenarycurve}
\end{equation}
where $r\in [-s, s]$ is the parameter of the curve; the variable~$s$ is equal to half of the span of the catenary (see Fig. \ref{fig:Intro}); and the value $a\in \mathbb{R}_{\geq 0}$  can be obtained by using the equation of the length of the catenary as
\begin{equation}
     \frac{\ell}{2} =
    a \; \sinh \left ( \frac{{s}}{a} \right ).
    \label{eq:sinh}
\end{equation}
Here we know the length of the cable $\ell$ and the distance $s$ that comes from the location of the quadrotors.
Since, this equation is transcendental, meaning that it is not possible to solve $a$ analytically, we will have to use a numerical solution.
In the catenary frame, the lowest point of the curve is $\boldsymbol{\alpha}(0)= \mathbf{0}$ in $\{\mathcal{C}\}$, and the location of the robots are
  $\boldsymbol{\alpha}(-s)= ^C\mathbf{x}_A,$ and 
  $\boldsymbol{\alpha}(s)= ^C\mathbf{x}_B.$


Each quadrotor has four propellers that generate a total thrust $f_i$ and a torque vector $\boldsymbol{\tau}_i$.
The translational and rotational dynamics of each quadrotor $i\in\{A,B\}$ are described by the Newton-Euler equations,
\begin{eqnarray}    %
    m {\mathbf{\ddot{x}}}_{i} &=&
    -m g\, \mathbf{e}_3+ {}^W\mathbf{R}_i {f}_i \mathbf{e}_3 + ^W\mathbf{R}_C\mathbf{t}_i ,\label{eq:newton}\\ 
    %
    %
    \mathbf{J}\dot{\boldsymbol{\omega}}_i &=&
    -\boldsymbol{\omega}_i \times \mathbf{J}\boldsymbol{\omega}_i 
    + \boldsymbol{\tau}_i,
    \label{eq:euler}
\end{eqnarray}
where $g$ is the gravity constant, $e_3 = [0,0,1]^\top$,   $\boldsymbol\omega_i$ is the angular velocity, and the vector $\mathbf{t}_i$ is the tension force that the cable generates on the $i$th quadrotor.

Using a geometric controller \cite{lee2010}, we can drive the robots to a desired attitude $^W\mathbf{R}_i^d$, and thrust $f_i^d$. In this way, the thrust in the $z$-axis of the quadrotor can be used to pull the cable in any direction. The force vector in $\{\mathcal{W}\}$ generated by each quadrotor is
$$
\mathbf{f}_i = f_i \:^W\mathbf{R}_i\, \mathbf{e}_3,
$$
the thrust vector is obtained by multiplying the rotation matrix of its attitude, i.e. $^W \mathbf{R}_A$, and the thrust generated by the motors, i.e. $f_A$, in the $z$-axis of the body frame.


The configuration space of the catenary curve is associated with its position, orientation, and span.
The sag depends on the length of the cable and the span, but the length is fixed, making the sag a function of the span.
Each quadrotor offers four inputs, and the catenary curve is described using five variables, satisfying the condition of the system to be fully actuated.
The objective of this work is to control the catenary in its configuration space. 



\section{Trajectory Tracking and Control}
\label{Sec:Trajectory}
\begin{figure*}[t]
    \centering
    {\includegraphics[width=0.8\linewidth]{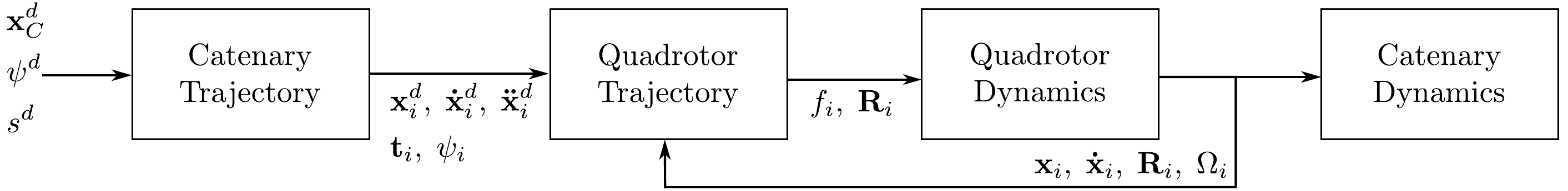}
    \caption{Control architecture for the catenary robot.
    }
    \label{fig:diagram}}
\end{figure*}

Given a desired trajectory for the catenary robot, specified by its reference point, 
$\mathbf{x}_C(t)$, orientation $\psi(t),$ and span $s(t)$, we design a controller to track the trajectory using the control inputs of the quadrotors $f_i$ and $ ^W\mathbf{R}_i$, $i\in\{A,B\}$.
An overview of the control architecture is illustrated in Fig.~\ref{fig:diagram}.
The first block receives the trajectory of the catenary robot and converts it into trajectories for the position of each quadrotor. The second block tracks the trajectory based on the attitude of the quadrotors. Finally, an attitude controller in SO(3) in the third block.

%


\subsection{Trajectory of the quadrotors}
Our first step is to convert the trajectory of the catenary, defined by $\mathbf{x}_C(t)$, $\psi(t),$ and $s(t)$, into the trajectory of the quadrotors, $\mathbf{x}_{A}^d(t)$, and $\mathbf{x}_{B}^d(t)$, including its first, and second derivatives.
\modification{The desired location}, velocity and acceleration of each quadrotor $i\in\{A,B\}$ in the world frame with respect to the catenary frame are
\begin{eqnarray}
            {\mathbf{x}_{i}^d} &=&
             \mathbf{x}_C^d 
             + {}^W\mathbf{{R}_{C}}(\psi)
              {}^C \mathbf{x}_{i},\\
            {\mathbf{\dot{x}}_{i}^d} &=& 
            \mathbf{\dot{x}}_C^d
            + {}^W\mathbf{\dot{R}_{C}}(\psi)  \;  {}^C\mathbf{x}_{i}
            + {}^W\mathbf{R_{C}(\psi)}  \;  {}^C\mathbf{\dot{x}}_{i}, \\
           {\mathbf{\ddot{x}}_{i}^d} &=& 
            \mathbf{\ddot{x}}_C^d
             + {}^W \mathbf{\ddot{R}_{C}}(\psi)   \;  {}^C\mathbf{x}_{i}
            +{}^W \mathbf{R_{C}(\psi)}  \;  {}^C \mathbf{\ddot{x}}_{i} \nonumber
            \\ &&  + 2  {}^W\mathbf{\dot{R}_{C}(\psi)}  \;  {}^C \mathbf{\dot{x}}_{i}. 
 \end{eqnarray}
The point $^C \mathbf{x}_{A}$ and its derivatives can be computed using the span of the catenary that comes from the function $s(t)$. 
By evaluating the catenary function $\boldsymbol{\alpha}$ at $-s$, i.e., $\boldsymbol{\alpha}(-s)= ^C\mathbf{x}_A,$
we can obtain the point $^C \mathbf{x}_{A}$ and compute its derivatives 
\begin{align}
            {{}^C\mathbf{x}_{A}} &= 
            \left\lbrack 
            \begin{array}{c}
                \modification{0,}\\
                \modification{-s,}\\
                a \left ( \cosh \left(\frac{{s}}{a}\right) - 1 \right )
            \end{array}
            \right\rbrack, \\
            {{}^C\mathbf{\dot{x}}_{A}} &=  \left\lbrack 
            \begin{array}{c}
                \modification{0,}\\
                \modification{-{{\dot{s}}},}\\
                \dot{a} \left ( \cosh \left ( \frac{{s}}{a} \right) -1 \right )+ \left ( \dot{s} 
                - \frac{s \dot{a} }{a} \right ) \sinh \frac{{s}}{a}\\
            \end{array}\right\rbrack,
 \end{align}
\begin{align}
            {{}^C\mathbf{\ddot x}_{A}} &= 
            \left\lbrack 
            \begin{array}{c}
                \modification{0,}\\
                \modification{-{\ddot{s}},}\\
                2\dot a \left(\frac{\dot s}{a} - \frac{s \dot a}{a^2} \right) \sinh\left(\frac{s}{a}\right) \\
                + a \left( \frac{ \dot s}{a} - \frac{s \dot a}{a^2} \right)^2 \cosh\left(\frac{s}{a}\right)\\ 
                + a \left(\frac{2s \dot a^2}{a^3} - \frac{2\dot a \dot s}{a^2} - \frac{s \ddot a}{a^2} + \frac{\ddot s}{a} \right) \sinh\left(\frac{s}{a}\right) \\
                + \ddot a \left(\cosh\left(\frac{s}{a}\right) - 1\right)
            \end{array}
            \right\rbrack .
 \end{align}
 In a similar way, the position and derivatives of the point $^C \mathbf{x}_{B}$ are obtained by evaluating  \eqref{eq:catenarycurve} in $s$, i.e., $\boldsymbol{\alpha}(s)= ^C\mathbf{x}_B$.
 In these equations, we still need to compute the variable of the catenary $a$ and its derivatives. Derivating \eqref{eq:sinh},
\begin{align}
    0 =& \left ( \dot{s} 
    - \frac{s \dot{a}}{a} \right ) \cosh \left ( \frac{{s}}{a} \right )
    + \dot{a} \; \sinh \left ( \frac{{s}}{a}  \right ) ,
    \label{eq:dsinh}
    \\
    0 =&  
     a \left ( \frac{\dot{s}}{a}
    - \frac{s \dot{a}}{a^2} \right )^2 \sinh \left ( \frac{{s}}{a} \right )
    + 2\dot{a}\left ( \frac{\dot{s}}{a}
    - \frac{s \dot{a}}{a^2} \right ) \cosh \left ( \frac{{s}}{a} \right ) 
    \label{eq:ddsinh}\\
    \nonumber
    &+ \left ( \frac{2s \dot{a}^2}{a^2}
    - \frac{2\dot{a} \dot{s}}{a}
    - \frac{s \ddot{a}}{a} 
    + \ddot{s} \right ) 
    \cosh \left( \frac{{s}}{a} \right)
    + \ddot{a} \; \sinh \left ( \frac{{s}}{a}  \right ).
 \end{align}
The equations \eqref{eq:sinh}, \eqref{eq:dsinh} and \eqref{eq:ddsinh} are transcendental and we cannot analytically solve our variables of interest $a$ and its derivatives. 
However, we need those values to control the catenary, so in each control iteration, we compute them by solving these equations numerically using the bisection method \cite{burden19852}.
In order to reduce the torsion of the cable, we want to make the quadrotors to always point in the direction of the normal vector of the catenary plane. Therefore, $\psi_i = \psi^d.$
In this way, we convert the inputs of the catenary in trajectories for the quadrotors.

\subsection{Tracking controller}
Using the desired position, velocity and acceleration of each quadrotor,
we can compute the errors in the trajectory as
$$\mathbf{e}_p=\mathbf{x}_i^d - \mathbf{x}_i, \text{ and } 
\mathbf{e}_v=\mathbf{\dot x}_i^d - \mathbf{\dot x}_i.
$$
Using a geometric controller \cite{lee2010}, each quadrotor can be driven to generate a thrust force vector $\mathbf{f}_i$.
Our desired force vector drives the tracking errors to zero, and compensates for the gravity force and the tension generated by the cable
\begin{eqnarray}
    \mathbf{f}^{d}_i &=& \mathbf{K}_p \mathbf{e}_p
    + \mathbf{K}_v \mathbf{e}_v
    + m\mathbf{\ddot{x}}_i^d - mg\mathbf{e}_3  + {}^W \mathbf{R}_C \mathbf{t}_i,
    \label{eq:dforce}
u\end{eqnarray}
where $\mathbf{K}_p$ and $\mathbf{K}_v$ are positive proportional matrices \modification{that drive position} 
and velocity errors to zero as time increases. $\mathbf{t}_i$ is the tension in the catenary frame in $A$, $B$ \modificationtwo{ and its direction is tangent to the catenary curve. A well-known result from the catenary curve \cite{lockwood1967book} is that tension at any point is $\mathbf{t}=[0, \pm w \, a, w \, z]^\top$, where $w=m/\ell$ is weight of rope per unit length. Then, evaluating in the two ends, we obtain} 
 \begin{equation*}
 \mathbf{t}_A= \left[  0,\: - w\, a,\, w\, {}^Cz_A \right]^\top, 
 \mathbf{t}_B= \left[  0,\:  w\, a,\, w\, {}^Cz_B \right]^\top.\\ 
 \end{equation*}
 It is important to highlight that the tension increases with the mass of the cable; requiring to increase the tilting angle of the quadrotors to compensate  the tension.
 
In order to drive the robot to generate the force vector, we can control its attitude in SO(3). The desired rotation matrix is defined by the unitarian vectors
\begin{eqnarray}
     \mathbf{z}_i^d =  \frac{\mathbf{f}^{d}_i}{\|\mathbf{f}^{d}_i \|},\:
     \mathbf{y}_i^d &=& \frac{\mathbf{z}_i^d \times \mathbf{x_i}}{\|\mathbf{z}_i^d \times \mathbf{x_i}\|},\text{ and }
     \mathbf{x}_i^d = \frac{\mathbf{y}_i^d \times \mathbf{z}_i^d}{\|\mathbf{y}_i^d \times \mathbf{z}_i^d\|}, \nonumber
 \end{eqnarray}
where the vector $\mathbf{x}_i$ comes from the desired orientation of the quadrotor, which is the same as the catenary, i.e. $\mathbf{x}_i=\text{Rot}_z(\psi)\mathbf{e}_1$. Then the desired rotation matrix is
$$
^W\mathbf{R}_i^d = \left [  \mathbf{x}_i^d, \mathbf{y}_i^d, \mathbf{z}_i^d \right].
$$
The thrust is then computed as
$$
f_i = \mathbf{f}^d_i \cdot\: ^W\mathbf{R}_i^d \mathbf{e}_3.
$$
The proof of the stability properties of this controller to generate the desired force $\mathbf{f}^d_i$ can be found in \cite{lee2010}.
By generating the desired force in \eqref{eq:dforce}, we can track the desired position and compensate for the gravity force and the cable's tension.
Our desired points $\mathbf{x}_i^d$, $i\in\{A, B\}$, and their derivatives come from the numerical solution. Although there is an error associated with the approximation, the feedback controller can compensate for it during running time. The important part is that the error does not accumulate over time since the numerical approximation has to be performed in each control loop.



\section{Experiments}
To test the catenary robot, we designed four different experiments\footnote{The source code for simulations and actual robots is available at\\ \texttt{https://github.com/swarmslab/Catenary\_Robot}}. First, we track a trajectory where the catenary point $\mathbf{x}_C$ is driven to $\mathbf{x}_C^d$, and then it remains static while the span and yaw orientation oscillate.
We demonstrate that even when the robots are moving around, the lowest point will remain at the same location once that this point is reached. This experiment is performed in both simulation and actual robots. Additionally, we change the cables to demonstrate the effect of increasing the weight of the cable. When the weight is big enough it affects the motion of the catenary robot and it is necessary to consider it in the model so the thrust compensates the extra force.  Second, we define a trajectory of the catenary to move through obstacles that require accurate motion.
Even though we do not have direct measurements of the curve, we are able to track a trajectory, based on the catenary equations from Section \ref{Sec:Trajectory}. 
In our third experiment,
\modification{ we interact with and attach to an umbrella handle (an object with a naturally hook-shaped attach point) by controlling the lowest point o the catenary curve.} Finally, the last experiment considers the transportation of a hook-shaped object. We generate a trajectory that allows the catenary robot to self-attach to the object and lift it to take it to another place in the environment.

\subsection{Simulation: Varying span and yaw}
\label{Sec:Simulations}
\begin{figure}[t]
  \centering{
  \includegraphics[trim=1in 1in 1in 1in, width=0.5\linewidth]{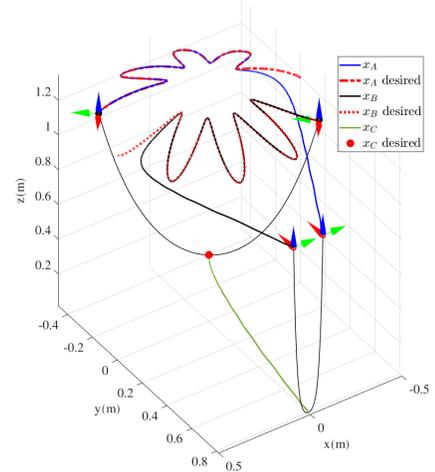}}
  \caption{3D trajectories of the quadrotors and the catenary tracking a span $s(t)$ and yaw $\psi(t)$ trajectories while $\mathbf{x}_C^d$ is static.}
  \label{fig:span_simulation}
\end{figure}

In this simulation,
we implemented the dynamic equations from \eqref{eq:newton} and \eqref{eq:euler}, and our controller in Matlab.
We tested with a trajectory that describes a flower-like shape by var
ying the span and yaw of the trajectory while maintaining the point $\mathbf{x}_C$ fixed.
We set the length of the cable as $\ell = 2\;m$, its mass as $m_C=7.6\; gr$, in the quadrotor mass $m =132 \; \text{g}$. The trajectory is defined by $\mathbf{x}_C^d = [0,0,0.4]$,  
$$\psi(t)=t/10\text{, and } s(t)=0.35 + 0.15\cos(t).$$
The result of the simulation is shown in Fig. \ref{fig:span_simulation}. The desired trajectories are denoted by the red dashed lines, the lowest point is in green solid line and the path of the quadrotors $\{A,B\}$ are in blue and black solid lines, respectively. 
It can be seen the catenary robot is able to lead the lowest point from $\mathbf{x}_C^0$ to $\mathbf{x}_C^d$ in finite time and remain there for the rest of the simulation. In the same way, the desired trajectory for the span is perfectly tracked by the controller.
Our next step is tracking the same trajectory with an actual catenary robot.

\subsection{Catenary robot design}
\label{Sec:Robot}
In our experimental testbed, we used the Crazyflie-ROS framework \cite{crazyflieROS} to command the robots. For the localization of the quadrotors, we use the motion capture system (Optitrack) operating at 120 Hz. The quadrotors internally measure their angular velocities using their IMU sensors. 
%
The original geometric controller in the firmware of the Crazyflie robot extended to include the tension of the cable. 
Since the original Crazyflie has a low payload ($<10$ g), we designed a quadrotor with brushless motors based on the crazybolt controller; its weight is $131.99$ g and its payload is $128$ g. 



\subsection{Experiments with actual robots}

The experiments for the catenary robot are the following.

\subsubsection*{\textbf{Experiment 1.1} Varying span and yaw}
\begin{figure}[t]
    \centering
    {$
    \begin{array}{cc}
         \includegraphics[width=0.36\textwidth]{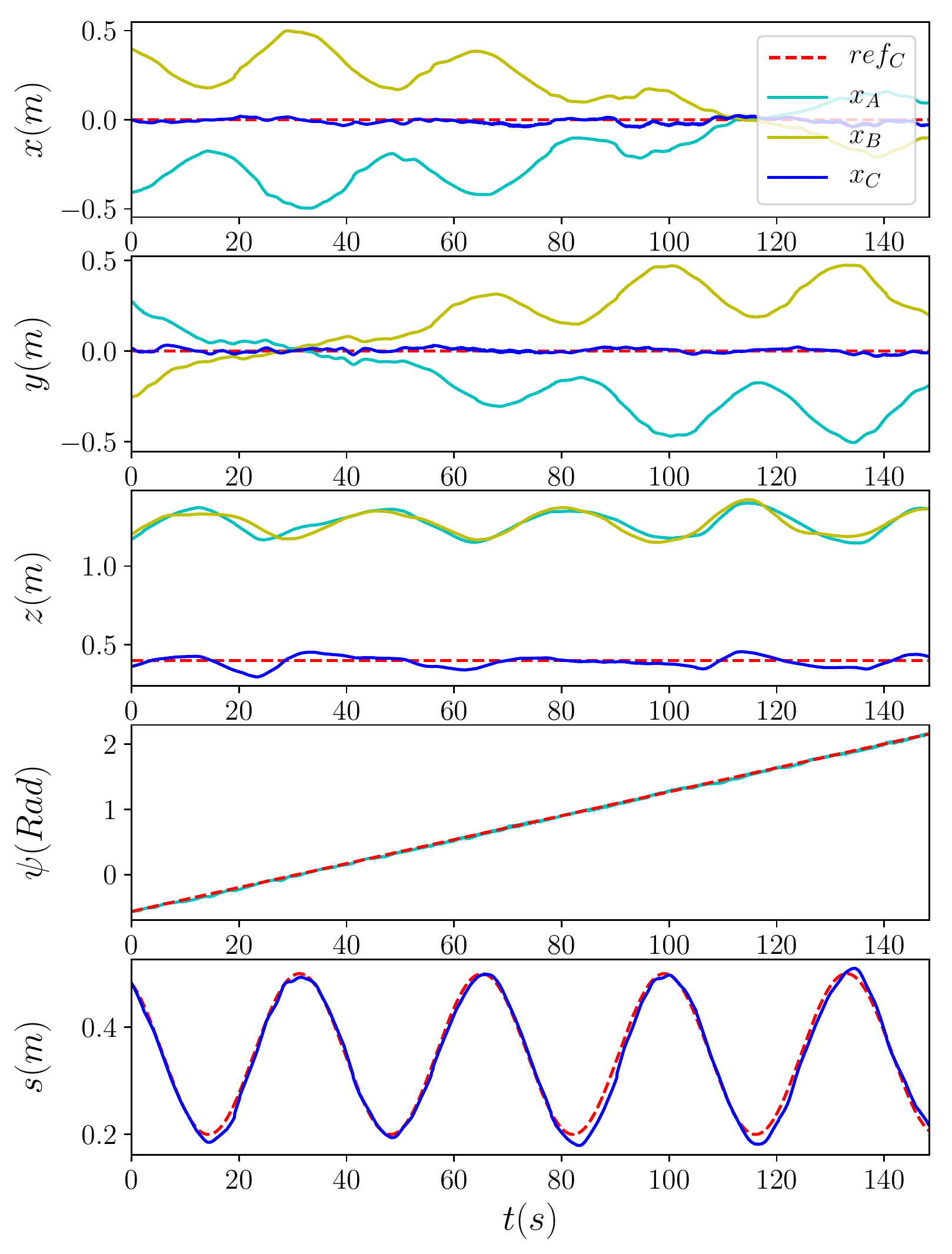}  \\
         \includegraphics[width=0.36\textwidth]{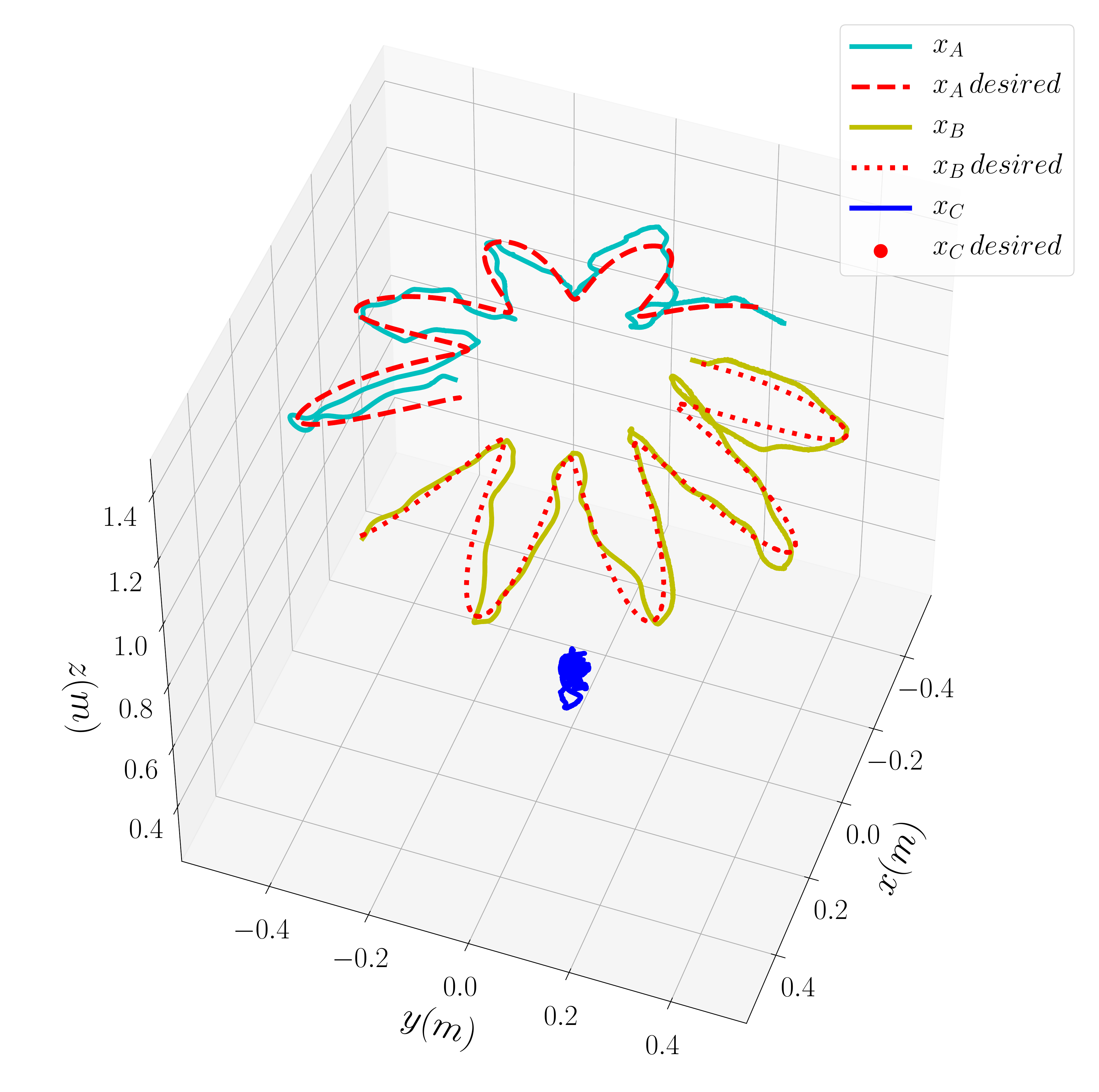} 
    \end{array}$
    \caption{Results of Experiment 1: the position of quadrotors $\mathbf{x}_A$, $\mathbf{x}_B$ and the  catenary point $\mathbf{x}_C$ with respect to each axis. Additionally the evolution of yaw-angle $\psi$ and span $s$ with their respecting desired trajectories that are in dashed red lines.}
    \label{fig:implementation_span}}
\end{figure}

\modification{
Performing the same experiment of the simulation in Fig. \ref{fig:span_simulation} but now using the actual robot. 
We present the results of the trajectory tracking 
in Fig. \ref{fig:implementation_span} for in $x$-, $y$-, and $z$-axis, as well as the yaw angle and span. Here is possible to see how the real implementation has a larger error in comparison with the simulation that assumes perfect conditions.}
It can be seen that there is an error while maintaining the lowest point of the catenary in a static location, which is close to zero in the $x$- and $y$-axis, but the $z$-axis has a small oscillating error. 
The average error in position is $\mu_x= 23e-3$, $\mu_y= 19e-3$, and $\mu_z= 90e-3$, and its standard deviations $\sigma_x=22e-3$, $\sigma_y=34e-3$, $\sigma_z=0.2$.
The errors for yaw-angle $\psi(t)$ and span $s(t)$ are $\mu_{\psi}= 10.7e-3 $, $\mu_s=3.5e-3$, and their standard deviation $\sigma_{\psi}= 6e-3$, and $\sigma_s= 0.17$. As a result, we can say that the $z$-coordinate and the span $s$ are the most sensitive variables.

\subsubsection*{\textbf{Experiment 1.2} Flying with different types of cables}
\modification{
We perform the previous experiment with different types of cables.  A rope,  a steel cable, and a plastic chain with a weight of 6.23, 14.17, and 56.39 g respectively. For the first and second cable, their weight is so low that the tension that it generates is neglectable. For the third cable, it is necessary to include tension term $\mathbf{t}_i$ in the controller for stabilizing the flight.
As a result, the quadrotors will fly tilting outwards to compensate the tension from the cable.
}

\subsubsection*{\textbf{Experiment 2.} Trajectory tracking}

\begin{figure}[t]
    \centering
    {\includegraphics[width=0.38\textwidth]{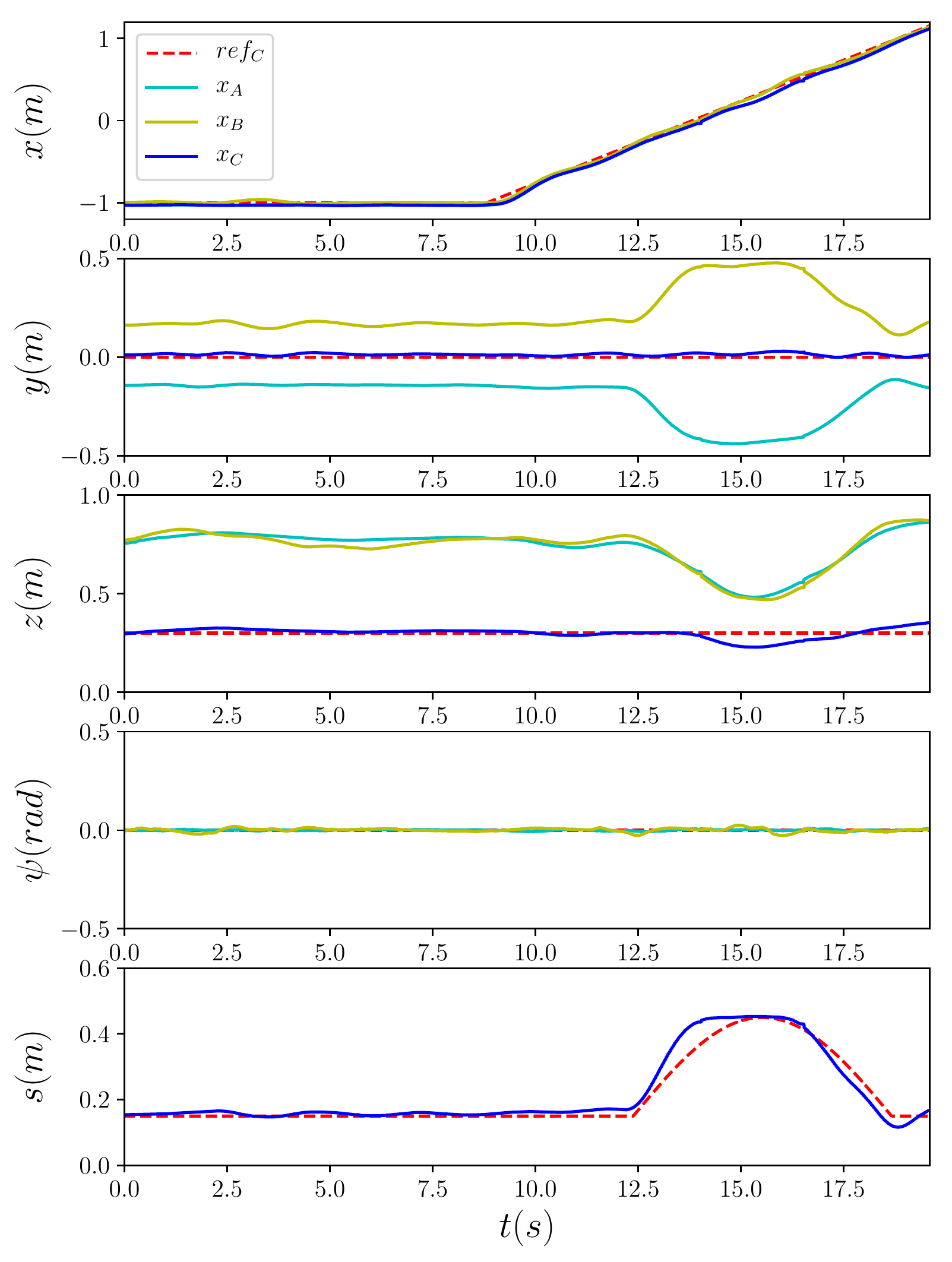}
    \caption{Results of Experiment 2: position of quadrotors $\{A,B\}$, $\mathbf{x}_A$, $\mathbf{x}_B$, and the lowest point of the catenary $\mathbf{x}_C$ are shown in light blue, olive, and blue respectively. All desired trajectories are in dashed red lines. 
    We show the signal in $x$-axis, $y$-axis, $z$-axis, yaw-angle $\psi$, and $s$ span of the catenary.}
    \label{fig:implementation_miminimum_trajectory}}
\end{figure}

In this experiment, the goal is to maintain a constant altitude of the lowest catenary point while moving in the $x$-axis. The trajectory is  $\mathbf{x}_c^d(t)=[t,0,0.3]$, $\psi^d= 0$, and
\begin{eqnarray}
    s(t) &&= \left \{ 
    \begin{array}{cc}
         0.3 & 0 \leq t < 4\pi   \\
         0.3 + 0.6 \sin(t) & 4\pi \leq t < 5\pi\\
         0.3 & 5\pi \leq t < \infty
    \end{array} \right.
    .
\end{eqnarray}
As shown in Fig. \ref{fig:implementation_miminimum_trajectory}, the lowest point of the catenary starts to follow the trajectory having an error in the $y$-axis and $x$-axis close to zero. When the point starts to change as expected by $\mathbf{x}_c^d(t)$, the quadrotors change the position making the span greater to maintain the lowest point at the same altitude. 
The average error in position is $\mu_x= 0.04 $, $\mu_y= 19e-3$, and $\mu_z= 0.8$, and its standard deviations $\sigma_x= 58e-3$, $\sigma_y= 28e-3$, $\sigma_z= 0.25$.
Similar to the previous experiment $z$ and $s$ are the most sensitive variables and they are mainly affected by the variation in span \modification{after 12.5 seconds.} 

\subsubsection*{\textbf{Experiment 3. }Pulling an umbrella}
\begin{figure}[t]
    \centering
    {\includegraphics[width=0.38\textwidth]{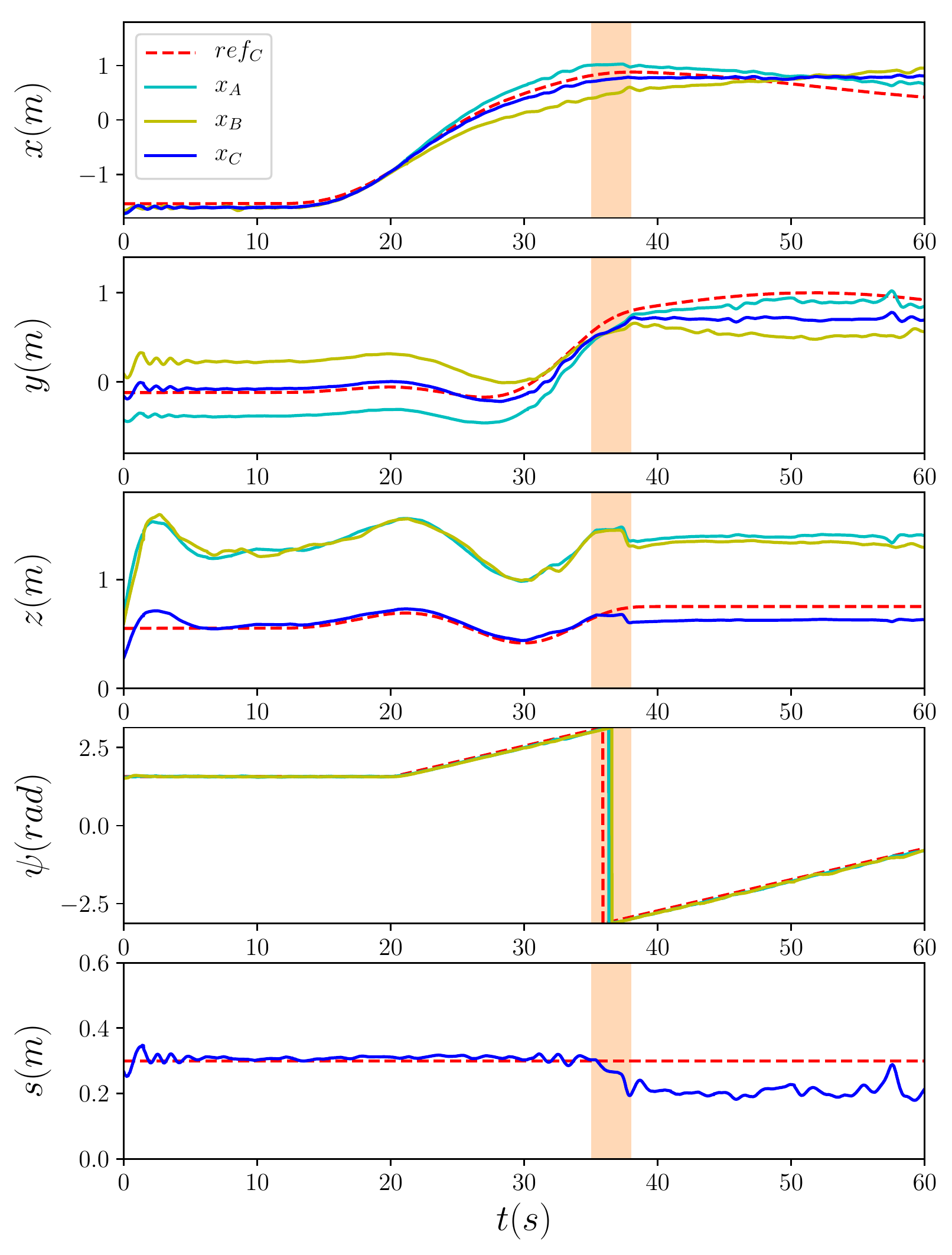}
    \caption{Results of Experiment 3: Attaching to the handle of an umbrella. the position of quadrotors $\mathbf{x}_A$, $\mathbf{x}_B$ and the  catenary point $\mathbf{x}_C$ with respect to each axis. Additionally the evolution of yaw-angle $\psi$ and span $s$ with their respecting desired trajectories that are in dashed red lines.}
    \label{fig:implementation_umbrella}}
\end{figure}

We want to use the catenary robot to interact with objects. By driving the robot in the proper manner, it can be used to pull objects with hook-like shapes.
In this case, we design a trajectory that allows the catenary robot to pull the handle of an umbrella. The procedure has three stages. First, the robot takes off to be placed in an initial location above ground. Second, the robot moves from the initial location to the front of the umbrella. Third, the lowest point of the catenary is right bellow the hook of the umbrella handle and then the catenary robot pulls the umbrella. 
We generate the minimum snap trajectory \cite{mellinger2011} for following waypoints $([[-1.6,-0.1,0.6],$ $[0.0,-0.2, 0.6],$ $[0.6, 0.17, 0.509],$ $[0.8,0.7,1.]])$ and fixing the desired  span at $s=0.3$.
\modification{It is noticeable that the catenary
robot is able to follow the generated trajectories by following the
waypoints. This is illustrated in Fig. \ref{fig:implementation_umbrella} through the evolution in time
of the position} of quadrotors $\mathbf{x}_A$, $\mathbf{x}_B$, catenary point $\mathbf{x}_C$, span $s$, and yaw-angle $\psi$. The average error in position is $\mu_x= 17e-3 $, $\mu_y= 92e-3$, and $\mu_z= 0.59$, and its standard deviations $\sigma_x= 1.06$, $\sigma_y= 0.37$, $\sigma_z= 0.2$. The errors for yaw-angle $\psi(t)$ and span $s(t)$ $\mu_{\psi}= 6e-3$, $\mu_s=28e-3$, and their standard deviation $\sigma_{\psi}= 1.9$, and $\sigma_s= 0.50e-3$. 
Notice that around time $35$ s the catenary robot starts to slightly deviate from the desired trajectory, this is given by the extra tension generated by the umbrella which is not considered in the model.

\subsubsection*{\textbf{Experiment 4. }Transporting a hook-shaped object}

We interact with an object that has a hook-shape.  The catenary robot
follows a trajectory that allows it to pull the object, lift it, and place it in a different location. \modification{Here is important to note that it is necessary to include the mass of the object in the model to generate the proper control input that accomplishes the transportation task.}



\section{Conclusions and Future Work}
In this work, we proposed a  robotic system, called the catenary robot, composed of a cable propelled by two quadrotors. We designed a controller to track trajectories for the five degrees of freedom of the catenary: position in three dimensions, yaw orientation, and span. Each degree of freedom can be controlled independently based on the forces generated by the quadrotors.
By estimating and controlling the catenary, we showed that it is possible to interact with objects that have hook-like shapes, e.g., an umbrella that has a hook shape in its handle. \modification{ We have demonstrated the successful functionality of our system}
in simulation and actual robots. 
In a future work, we want to use the catenary robot to manipulate object without hook-shapes. 



\bibliographystyle{ieeetr}
\bibliography{referencias.bib}

\end{document}